\pdfoutput=1
\documentclass[11pt]{article}

\usepackage[preprint]{neurips_2023}

\usepackage{times}
\usepackage{latexsym}
\usepackage{hyperref}
\usepackage{arydshln}

\usepackage[T1]{fontenc}
\usepackage[utf8]{inputenc}
\usepackage{amsfonts}
\usepackage{booktabs}
\usepackage{amsmath}
\usepackage{graphicx}
\usepackage{xcolor} 
\usepackage[most]{tcolorbox}

\usepackage{microtype}

\definecolor{backquote}{RGB}{249,245,233}
\newtcolorbox{myquote}[1][]{%
    enhanced, breakable, 
    size=minimal,
    frame hidden, boxrule=0pt,
    sharp corners, top=2pt,
    colback=backquote,
    #1
}

\title{Exploring the impact of low-rank adaptation on the \\ performance, efficiency, and regularization of RLHF}

\author{Simeng Sun \hspace{3mm} Dhawal Gupta
  \hspace{3mm}  Mohit Iyyer \\
University of Massachusetts Amherst \\
 \texttt{\{simengsun,dgupta,miyyer\}@cs.umass.edu} \\
}

\begin{document}
\maketitle

\begin{abstract}
During the last stage of RLHF, a large language model is aligned to human intents via PPO training, a process that generally requires large-scale computational resources. In this technical report, we empirically investigate an efficient implementation of RLHF using low-rank adaptation (LoRA), which allows us to align the LLaMA 7B checkpoint on the Alpaca dataset~\citep{alpaca} using only two A100 GPUs instead of the eight required for full model fine-tuning. Despite tuning only 0.2\% of LLaMA 7B's parameters, our implementation achieves \emph{better} performance than the publicly-released AlpacaFarm checkpoint~\citep{dubois2023alpacafarm} with full model fine-tuning. 
Next, we analyze several configurations of our LoRA-based PPO implementation, varying the form of the KL regularization term in the training objective. We find that (1) removing this penalty term does not harm performance on the AlpacaFarm evaluation set under our LoRA setup; (2) other regularizers, such as Jensen-Shannon divergence, lead to improved performance;  and (3) while PPO training negatively impacts the factuality of model-generated responses, training with LoRA largely mitigates this effect. We release our code and pretrained checkpoints to facilitate future research on more efficient RLHF.\footnote{\url{https://github.com/SimengSun/alpaca_farm_lora}}
\end{abstract}
\section{Introduction}

Reinforcement learning from human feedback~\citep[][RLHF]{ouyang2022training} is a technique used to align large language models (LLMs) with the intents of human users. While companies such as OpenAI, Google, and Anthropic provide blackbox access to LLMs tuned with RLHF, their codebases remain proprietary. While the development of open-source  libraries such as TRL~\citep{vonwerra2022trl}, TRLX~\citep{trlx-library}, and AlpacaFarm~\citep{dubois2023alpacafarm} has enabled other researchers to experiment with RLHF, the required experimental settings are computationally infeasible for most under-resourced labs. For example, aligning the pre-trained LLaMA 7B checkpoint~\citep{touvron2023llama} with RLHF requires \emph{eight} Nvidia A100 80GB GPUs using the AlpacaFarm library. Currently, a single node with 8 A100 GPUs costs around \$200K to purchase,\footnote{For example, see this \href{https://shop.lambdalabs.com/deep-learning/servers/hyperplane/customize}{quote} from Lambda Labs.} and renting such resources from cloud providers borders on the impossible due to limited supply and huge demand. 

\paragraph{Why does RLHF require so many GPUs?} RLHF involves three stages that each requires fine-tuning a pre-trained LLM: (1) supervised fine-tuning on instruction following data; (2) reward model training on human preference data; and (3) fine-tuning via proximal policy optimization~\citep[][PPO]{schulman2017proximal}. The third stage, in which the RLHF-aligned LLM is actually created, is also the most expensive because it requires storing multiple large models (e.g., policy, value, reward, reference policy) along with gradients and optimizer states in GPU memory. The AlpacaFarm library leverages fully sharded data parallel training~\citep[][FSDP]{zhao2023pytorch} to distribute the parameters of these models across multiple GPUs, which along with flash attention~\citep{dao2022flashattention} enables processing two examples per GPU.  

\paragraph{Saving memory with LoRA:}
Low-rank adaptation~\citep[][LoRA]{hu2021lora} is a parameter-efficient method to fine-tune large language models. In LoRA, the pre-trained LLM is frozen while only low-rank decomposition of the weight matrices (commonly just the projection matrices in self-attention) are optimized. The rank of the decomposition matrices is typically very small (e.g., 8 or 64) compared to the dimensionality of the hidden states (e.g., 4096), which greatly reduces memory consumption.
\begin{myquote}[colback=yellow!20]
\emph{Contribution \#1:} In this technical report, we implement the third and most resource-demanding stage of RLHF (PPO) with LoRA and manage to bring down the hardware requirements from eight to two A100s. We observe no performance degradation when using LoRA; in fact, with just 10 hours of LoRA-based PPO training, we outperform the publicly-released AlpacaFarm checkpoint (trained via full model fine-tuning) in terms of win rate against \texttt{text-davinci-003}. To facilitate reproducibility and future research, we release our code, which modifies the AlpacaFarm library to support LoRA and alternative regularization schemes, as well as our LoRA weights for LLaMA-7B. All experiments in this report were conducted on two A100 80GB GPUs. 
\end{myquote}

\paragraph{KL regularization may not be critical when using LoRA.} The third stage of RLHF (i.e., PPO optimization) attempts to maximize the expected reward of the policy model while also penalizing large deviations from the pre-trained base model. This penalty is implemented via an approximation of KL divergence~\citep{kullback1951information} between the policy and  pre-trained reference LLM. 

\begin{myquote}[colback=yellow!20]
\emph{Contribution \#2:} While prior work has considered this penalty critical to the successful application of RLHF~\citep{ouyang2022training}, our experiments show that the KL penalty can be completely removed when using LoRA without lowering the resulting model's win rate. Additionally, we discover that implementing this penalty using different divergence estimators (e.g., Jensen-Shannon divergence) can lead to higher win rates on the AlpacaFarm evaluation set.
We hypothesize that LoRA itself acts as a powerful regularizer since most of the pretrained LLM's parameters (e.g., in the feed-forward layers) are left unchanged, and as such additional regularization is not as critical as in full model fine-tuning. This can lead to further memory reductions, as it may not be necessary to keep the reference policy in memory when using LoRA. 
\end{myquote}
 
\begin{myquote}[colback=red!12]
\emph{Limitations of this work:} Due to limited compute (i.e., no access to a node with 8 A100 80GB GPUs), we could not run RLHF on the LLaMA 7B checkpoint using full model fine-tuning. Thus, it is infeasible for us to assess the impact of removing or replacing the KL penalty term on full model fine-tuning. These experiments are critical to verify our hypothesis about LoRA's regularization effect. Additionally, because FSDP implementations did not support disabling gradients for specific parameters at the time of our experiments, we do not use any model parallelism during training, which makes it difficult to experiment with the larger LLaMA checkpoints. In our current codebase, the policy and value models are stored on one GPU while the reference policy and reward model are stored on the other; the latter GPU is thus only sparsely utilized. Finally, as we only experiment with the AlpacaFarm data, our conclusions may not generalize to other domains, languages, or to more complex instruction sets.
\end{myquote}

\section{Implementing RLHF with LoRA} \label{sec:method}

To ground our discussion, we first provide a brief overview of the  experimental settings that we consider in our work, which builds on the AlpacaFarm platform~\citep{dubois2023alpacafarm}. Then, we discuss how we incorporate LoRA into the PPO step of RLHF, which leads to slightly \emph{improved} performance while also cutting down on memory consumption (from eight to two A100s). 

\subsection{Experimental settings}

All of our experiments are conducted on the publicly-available LLaMA 7B checkpoint, which was pre-trained on 1 trillion tokens using 82K GPU hours (A100 80GB)~\citep{touvron2023llama}. 
We build our codebase on top of the open-source AlpacaFarm platform~\citep{dubois2023alpacafarm} instead of other open-source RLHF libraries  because AlpacaFarm also implements rigorous evaluations of RLHF-tuned instruction-following models.

\paragraph{RLHF data:} We follow~\citet{dubois2023alpacafarm} by applying RLHF to the pre-trained LLaMA model using the Alpaca-52K instruction-following dataset~\citep{alpaca}. Each example in this dataset, which was automatically generated using \texttt{text-davinci-003}
, contains an instruction for some task and a corresponding demonstration (i.e., an optional input and output for that task). The dataset contains a diverse set of tasks spanning both open-ended instructions (e.g., \emph{Describe the impact of the coronavirus on the economy}) and highly constrained ones (\emph{Standardize the following date to the ISO 8601 format}). Each of the three stages of RLHF is performed using a different split of the Alpaca-52K data: supervised fine-tuning with 10K examples, reward modeling with 10K instructions / model-generated outputs for which human raters provided preference judgments, and PPO with 20K instructions.\footnote{The rest of the data is left unused.}

\paragraph{Evaluation by win rate:} AlpacaFarm collates existing open-source instruction-following datasets~\citep{wang-etal-2023-self-instruct,bai2022training,vicuna2023,koala_blogpost_2023} to form an evaluation dataset of 805 diverse instructions. 
To evaluate two models against each other on this dataset, we first collect responses generated by both models for all 805 instructions. Then, a pool of large language models (\texttt{GPT-4-0314}, \texttt{GPT-3.5-turbo}, \texttt{text-davinci-003}) are prompted to provide preference judgments (i.e., which model's output is better for a given instruction) by simulating human annotators, which allows us to compute the \emph{win rate} of one model over the other. The LLM-based simulated workflows were shown to highly correlate with human raters in system-level comparison with a Spearman correlation of 0.98~\citet{dubois2023alpacafarm}.
As in the original codebase, we compute the win rate of each of our model configurations against OpenAI's \texttt{text-davinci-003}.

\paragraph{Baseline model:}
\citet{dubois2023alpacafarm} release checkpoints for the SFT-10K step (stage 1 of RLHF), the reward model trained on human preference (stage 2), and the PPO-optimized stage 3 checkpoint. Their SFT-10K model\footnote{\url{https://huggingface.co/tatsu-lab/alpaca-farm-sft10k-wdiff}} reaches a win rate of 37\% against \texttt{text-davinci-003}, while the PPO-optimized checkpoint\footnote{\url{https://huggingface.co/tatsu-lab/alpaca-farm-ppo-human-wdiff}} has a win rate of 47\%. 

In this technical report, we use the publicly-released AlpacaFarm checkpoints for the SFT and reward modeling stages of RLHF. These first two stages are not as memory intensive as the third stage,\footnote{Full-model fine-tuning for the first two stages requires 4 A100 80GB GPUs; however, only one A100 is needed if using LoRA.} and prior work~\citep{hu2021lora,santacroce2023efficient} has shown that LoRA is very effective for fine-tuning LLMs with limited resources. Our focus is on the third stage of RLHF (PPO), in which these previous two checkpoints have to be kept in memory along with the policy network, where we perform PPO with LoRA using the third split of 20K instructions.

 \subsection{Reducing RLHF's memory consumption with LoRA}
To alleviate the memory consumption of the third stage of RLHF, we employ low-rank adaptation~\citep{hu2021lora} to align the model. 
In LoRA, an input hidden state $\mathbf{h}_{\text{in}} \in \mathbb{R}^{d}$ is projected to $\mathbf{h}_{\text{out}} \in \mathbb{R}^{d}$ via a weight matrix $\mathbf{W}$ $\in \mathbb{R}^{d\times d}$ and two low-rank decomposition matrices $\mathbf{A} \in \mathbb{R}^{k \times d}$ and $\mathbf{B} \in \mathbb{R}^{d \times k}$, where $k$ is the rank of the low-rank matrices ($k \ll d$) and $\alpha$ is a scaling hyperparameter:
\begin{equation}
    \mathbf{h}_{\text{out}} = (\mathbf{W} + \frac{\alpha}{k}\mathbf{B}\mathbf{A})\mathbf{h}_{\text{in}}
\end{equation}
During LoRA training, $\mathbf{W}$ is kept frozen while the decomposition matrices $\mathbf{A}$ and $\mathbf{B}$ are trained. 

\paragraph{Experiment details:} We follow the original LoRA setup of~\citet{hu2021lora} by setting the rank $k$ to $8$. Additionally, we set the scaling hyperparameter $\alpha$ to 64, which is a critical decision: using the default value of $\alpha=1$ in the LoRA codebase\footnote{\url{https://github.com/microsoft/LoRA/tree/main}} reduces the win rate by $\sim$ 6 points compared to when $\alpha=64$. We apply LoRA only to the projection matrices (key, query, value, and output) in the attention layers. We also apply normal fine-tuning to the final projection head of size $[4096, 1]$ in the value function (critic), which leads to faster convergence in our experiments. Finally, we add dropout to all tuned layers with $p=0.1$.  
In total, we optimize 16.7M parameters  with LoRA ($\sim0.2\%$ of LLaMA 7B's parameters). For all experiments, we set the KL coefficient $\beta=0.02$ (see Eq.~\ref{eq:1}), the rollout batch size  to 256, and the gradient update batch size to 128. We perform three training runs for each of our configurations and report mean and standard error with bootstrap sampling. 

\subsection{LoRA is effective for PPO training}
\begin{figure*}
    \centering
    \includegraphics[width=0.5\textwidth]{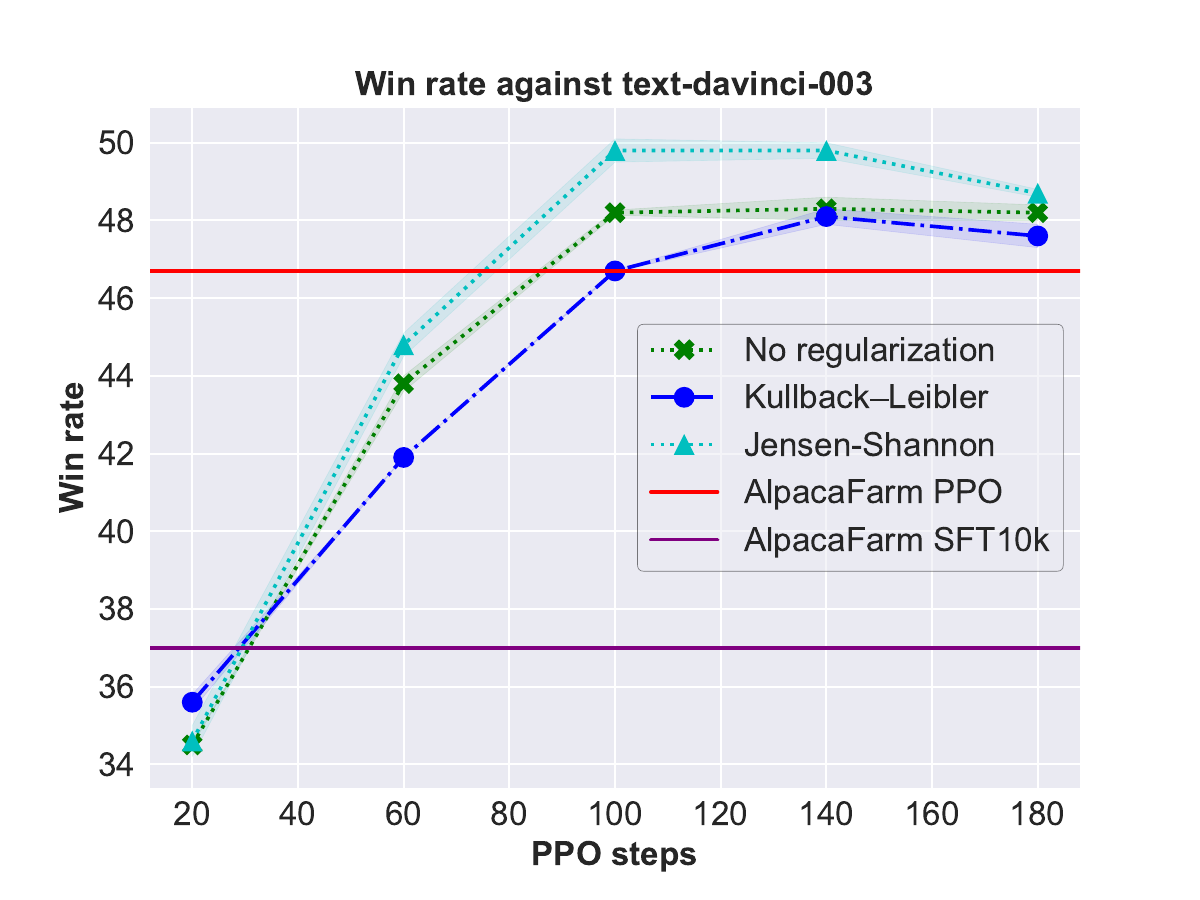}
    \caption{Win rate against \texttt{text-davinci-003} on the AlpacaFarm evaluation set plotted as a function of PPO steps. The Jensen-Shannon estimator consistently outperforms no regularization, which in turn outperforms the standard KL estimator. We observe win rates plateauing after roughly 100 steps.}
    \label{fig:win_rates}
\end{figure*}

Our experiments in Table~\ref{tab:kl_terms} show that LoRA is a powerful parameter-efficient adaptation method for PPO training. While the publicly-released PPO-optimized checkpoint of~\citet{dubois2023alpacafarm} reaches a win rate of 46.7\% with full model fine-tuning,\footnote{We download their checkpoint and rerun the evaluation ourselves using pooled LLM annotators. While the AlpacaFarm leaderboard was recently updated to report a higher win rate of 49.25, our evaluation result closely matches their previously-reported number of 46.6 when using the pooled annotators before the update on Jun 23rd, 2023. (see here: ~\url{https://github.com/tatsu-lab/alpaca_farm/blob/0cfd0bce0506b3d68998c82b17a160d7da1d99a0/src/alpaca_farm/auto_annotations/eval.py\#L23}).} our corresponding LoRA model (third row of Table~\ref{tab:kl_terms}) obtains a slightly \emph{higher} win rate of 47.5\% despite optimizing just a small fraction of the parameters. Note that this is as controlled of an experiment as we can perform: we use the same KL regularization penalty as that of the published checkpoint and also the same batch size (16). We report win rate after 100 steps of training,\footnote{We report results after 100 steps because we empirically observe the win rate of most configurations plateaus after this point (see Figure~\ref{fig:win_rates} for win rates at other points during training).} which takes around 10 hours of training on two A100s. The released AlpacaFarm PPO checkpoint was trained for 20 PPO steps,\footnote{Verified after email communication with one of the authors.} which is equivalent to 40 steps in our experimental setup. ~\citet{dubois2023alpacafarm} noted that optimal PPO performance is achieved between 20 to 80 PPO training steps under their setup, and they perform model selection based on simulated win rates.\footnote{\url{https://github.com/tatsu-lab/alpaca_farm/issues/60}} In contrast, we observe increases in win rate until $\sim$100 steps in our LoRA setup (Figure~\ref{fig:win_rates}). Having established the effectiveness of LoRA for RLHF, we now turn to investigating the regularization effects that it has on PPO training. 

\begin{table*}[]
    \centering
    \footnotesize
    \scalebox{0.89}{
    \begin{tabular}{lllll}
    \toprule
       \parbox{2cm}{\textbf{Adaptation}\\\textbf{method}}  &  \textbf{Regularization} & \textbf{Divergence estimator} & \textbf{Win rate} \\ \midrule
       \multicolumn{5}{l}{\footnotesize{\emph{Publicly-released supervised fine-tuning (SFT-10k) checkpoint from~\citet{dubois2023alpacafarm}}}}\vspace{0.1cm} \\
       \hspace{1em} Full model tuning & -  & - & \hspace{1em} $37.0$ \vspace{0.5em}\\\midrule
        \multicolumn{5}{l}{\footnotesize{\emph{Publicly-released PPO checkpoint from~\citet{dubois2023alpacafarm}}}}\vspace{0.1cm} \\
       \hspace{1em} Full model tuning & Clamped KL & $\max(0, \log \pi_{\theta}(y) - \log \pi_{\text{ref}}(y))$ & \hspace{1em} $46.7$ \vspace{0.5em}\\
       \midrule
       \multicolumn{5}{l}{\footnotesize{\emph{Our PPO models trained with LoRA}}}\vspace{0.1cm} \\
       \hspace{1em} LoRA & Clamped KL  & $\max(0, \log \pi_{\theta}(y) - \log \pi_{\text{ref}}(y))$ & \hspace{1em} $47.5 \pm 0.2$ \vspace{0.5em}\\ 
       \hspace{1em} LoRA &  KL   & $\log \pi_{\theta}(y) - \log \pi_{\text{ref}}(y)$ & \hspace{1em}  $46.7\pm 0.02$ \vspace{0.5em}\\
       \hspace{1em} LoRA &  Bregman  & $\frac{\pi_{\text{ref}}(y)}{\pi_{\theta}(y)} - 1 - \log \frac{\pi_{\text{ref}}(y)}{\pi_{\theta}(y)}$ & \hspace{1em}  $49.0 \pm 0.1$ \vspace{0.5em}\\
       \hspace{1em} LoRA &  Squared error   & $\frac{1}{2}(\log \pi_{\theta}(y) - \log \pi_{\text{ref}}(y))^2$ & \hspace{1em} $47.1 \pm 0.1$ \vspace{0.5em}\\
       \hspace{1em} LoRA &  Jensen-Shannon  & $\frac{1}{2} \max(0, \log \pi_{\theta}(y) - \log \frac{1}{2}(\pi_{\theta}(y) + \pi_{\text{ref}}(y))) + $ & \hspace{1em} $49.8 \pm 0.3$\vspace{0.2em} \\
       & &  $\frac{1}{2} \max(0, \log \pi_{\text{ref}}(y) - \log \frac{1}{2}(\pi_{\theta}(y) + \pi_{\text{ref}}(y)))$ & \vspace{0.5em}\\
       \hspace{1em} LoRA &  \emph{No regularization} &  - & \hspace{1em} $48.2 \pm 0.1$ \vspace{0.5em}\\
       \hspace{1em} LoRA &  Dropout only ($p$=0.5) &  - & \hspace{1em} $48.4 \pm 0.2$ \vspace{0.5em}\\
       \bottomrule
    \end{tabular}}
    \caption{Our experiments demonstrate that  (1) LoRA is a competitive parameter adaptation method to full-model fine-tuning for PPO training in RLHF; and (2) PPO training with LoRA does not require any additional regularization (KL or otherwise) to succeed. We evaluate the win rate of each model in the table against \texttt{text-davinci-003} on the AlpacaFarm evaluation data. Preference judgments to calculate win rate are simulated by a pool of automated LLM annotators (e.g., GPT-4). We evaluate the public AlpacaFarm checkpoints in the top two rows (SFT-10k and PPO); for each of the remaining rows, we perform three runs of PPO optimization with LoRA and the corresponding regularizer, and we report the mean and standard error with bootstrap sampling.
    }
    \label{tab:kl_terms}
\end{table*}

\section{KL regularization in PPO training}
First, we provide a brief overview of PPO training in the third stage of RLHF, highlighting the KL regularization term that is commonly regarded as critical for successful training. We experiment with different variants of this KL penalty with our LoRA setup (including no regularization at all), which we fully specify here.

\subsection{PPO training}
In the third stage of RLHF, the SFT model from the first stage (which is a model with basic instruction-following capabilities) is reinforced with PPO against the reward model trained in the second stage. 
PPO training~\citep{schulman2017proximal} iterates between a rollout phase and an  optimization phase. 

\textbf{Rollout}: In rollout, the policy  $\pi_{\theta}$ being optimized (i.e., the LM undergoing alignment) generates responses to a batch of input instructions. Each response is then assigned a scalar score by the reward model. This score is then used to estimate the advantage\footnote{A positive advantage suggests that when $\pi_{\theta}$ generates token $y$ given a certain prefix, it receives a reward higher than the average reward expected from generating other tokens in the vocabulary given the same prefix.
}
 for optimization in the next phase. 

\textbf{Optimization}: The policy is optimized to maximize a surrogate objective using the advantage estimated during rollout. Then, the newly-optimized policy is used to generate responses in the next rollout phase.
The surrogate objective in the optimization phase has the form
\begin{equation}
    \mathcal{J}(\theta) = \mathbb{E}[\min(r(\theta) \hat{A}_{\theta_{\text{old}}}, \text{clip}(r(\theta), 1-\epsilon, 1+\epsilon) \hat{A}_{\theta_{\text{old}}})],
    \label{eq:ppo}
\end{equation}
where $r(\theta) = \frac{\pi_{\theta}(y\mid x)}{\pi_{\theta_{\text{old}}}(y\mid x)}$ denotes the output probability ratio between the current policy $\pi_{\theta}$  and the policy $\pi_{\text{old}}$ from the previous rollout step, computed on responses $y$ sampled during the previous rollout. $\hat{A}$ denotes the estimated advantage~\citep{schulman2018highdimensional}, which depends on the reward value; we refer readers to other materials~\citep{schulman2018highdimensional,lilianblog} for understanding advantage estimation. The clipped output probability ratio stabilizes training by preventing large policy updates, thus discouraging $\pi_{\theta}$ from deviating too much from $\pi_{\text{old}}$ during optimization. 

\subsection{KL Regularization}

While the clipped ratio in Equation~\ref{eq:ppo}  constrains the extent to which $\pi_{\theta}$ can change from a recent policy $\pi_{\text{old}}$,  $\pi_{\theta}$ can still reach a sub-optimal region by ``reward hacking''~\citep{pan2022the} after enough rollout-optimization steps.  To avoid this, a KL regularization term is added to penalize $\pi_{\theta}$ when it deviates too far from a reference policy $\pi_{\text{ref}}$ during PPO training; the reference policy is usually set to the output of the first stage of RLHF (i.e., the SFT-10k checkpoint). 

Formally, let $\mathbf{x}$ be an instruction and $\mathbf{y}$ be a corresponding response of $L$ tokens sampled during rollout. While the reward model produces a scalar value $r(\mathbf{x}, \mathbf{y}) \in \mathbb{R}$ given the pair $(\mathbf{x},\mathbf{y})$, the total reward $\mathbf{r}(\mathbf{x}, \mathbf{y})$ of the response $\mathbf{y}$ is a vector of dimensionality $\mathbb{R}^{L}$ due to the KL penalty:
\begin{equation} \label{eq:1}
    \mathbf{r}(\mathbf{x}, \mathbf{y}) = \begin{bmatrix}0 & 0 & \dots & r(\mathbf{x}, \mathbf{y}) \end{bmatrix}^\top - \beta \text{KL}(\pi_{\theta}(\mathbf{y} \mid \mathbf{x} ) , \pi_{\text{ref}}(\mathbf{y} \mid \mathbf{x}))
\end{equation}
Here, $\text{KL}(\cdot, \cdot)$ denotes the KL divergence~\citep{kullback1951information} between  $\pi_{\theta}$ and $\pi_{\text{ref}}$ at each position (token) of $\mathbf{y}$, and the scalar reward $r$ is added to the last position of the KL term. The shaped reward $\mathbf{r}$ is then used to estimate the advantage, which is at the core of PPO. 

As it is memory-consuming to compute the  KL divergence over the entire vocabulary at each timestep,\footnote{
Computing the true KL requires summing over the entire vocabulary, which is not economical if done for every token in the sequence.
}
prior work approximates the KL term via Monte-Carlo estimation~\citep{schulman2017trust}. Does the form of this approximation make an impact when implementing RLHF with LoRA? In this report, we compare the following divergence estimators based on win rate:  
\begin{itemize}
    \item \textbf{KL divergence:} \hspace{0.5em} Given the distributions over the vocabulary predicted by both the current policy $\pi_{\theta}$ and the reference policy $\pi_{\text{ref}}$ at token $y$ of the response, the Kullback-Leibler divergence is defined as  $\text{KL}(\pi_{\theta}(y), \pi_{\text{ref}}(y)) = \sum_{y\in\mathcal{V}} \pi_{\theta}(y) \log \frac{\pi_{\theta}(y)}{\pi_{\text{ref}}(y)}$. By minimizing the KL divergence, we effectively add an entropy regularizer that diversifies the generated tokens while also maintaining high likelihood under the reference policy. In practice (and in many open-source implementations), the true KL divergence
    is approximated as $\log \pi_{\theta}(y) - \log \pi_{\text{ref}}(y)$, which saves memory as only the probability of the generated token $y$ is stored instead of the entire distribution. While this estimator is unbiased, it also suffers from high variance~\citep{schulmanblog}; to enforce non-negativity, the AlpacaFarm implementation clamps the minimum value to zero, which reduces variance at the cost of biasing the estimator.
    
    \item \textbf{Alternative divergence approximations:} \hspace{0.5em} ~\citet{schulmanblog} proposes two alternative KL approximations that have lower variance. The first takes the form of squared error between $\log \pi_{\theta}(y)$ and $\log \pi_{\text{ref}}(y)$, which is biased and approximates a different $f$-divergence measure.\footnote{$f(x):=\frac{1}{2}(\log x)^2$} The second one is the Bregman divergence $B_{\mathcal{F}}(\frac{\pi_{\text{ref}}(y)}{\pi_{\theta}(y)}, 1)$ associated with the convex function $\mathcal{F}(x):=-\log x$, which is an unbiased KL estimator.\footnote{
    It is an unbiased estimator of KL because
    \begin{align}
        \mathbb{E}_{y\sim \pi_{\theta}}[\frac{\pi_{\text{ref}}(y)}{\pi_{\theta}(y)} - 1 - \log \frac{\pi_{\text{ref}}(y)}{\pi_{\theta}(y)}] &= -1 + \sum_{y\in V} \pi_{\theta}(y)\frac{\pi_{\text{ref}}(y)}{\pi_{\theta}(y)} - \sum_{y\in V} \pi_{\theta}(y) \log \frac{\pi_{\text{ref}}(y)}{\pi_{\theta}(y)} \nonumber \\
        &= -1 + 1 + \text{KL}(\pi_{\theta}(y), \pi_{\text{ref}}(y)) \nonumber
    \end{align}
    }
    Both alternatives are shown  to reduce variance and bias in a toy setting by~\citet{schulmanblog}; in this report, we verify their effectiveness on real natural language tasks.
    \item \textbf{Jensen-Shannon divergence:}  \hspace{0.5em} We also experiment with the Jensen-Shannon divergence,
    which is the average of  the KL divergence for both $\pi_{\text{ref}}$ and $\pi_{\theta}$ against the intermediate distribution $\pi_m(y) = \frac{1}{2}(\pi_{\text{ref}}(y) + \pi_{\theta}(y))$. Note that the responses are sampled from $\pi_{\theta}$ instead of $\pi_{\text{ref}}$. Similar to the AlpacaFarm implementation, the minimum value is clamped to zero for stabilized training. We fully specify the estimator in Table~\ref{tab:kl_terms}.
    \item \textbf{No KL regularization:}  \hspace{0.5em} Finally, to evaluate whether KL regularization is necessary under the LoRA setup, we also discard KL penalty entirely in two of our configurations, one with dropout~\citep{JMLR:v15:srivastava14a} and the other without. 
\end{itemize}

\section{Results \& analysis}

We observe several interesting phenomena while varying the regularization estimator within our LoRA RLHF setup. First and perhaps most notably, completely \emph{removing} the KL regularization penalty does not affect win rate (it actually increases from 47.5\% to 48.2\% given same amount of compute). Of the alternative estimators, the Jensen-Shannon divergence yields the highest overall win rate (49.8\%). We also compare the quality of the estimator against the true KL divergence computed over the full distributions and discover that win rate is not necessarily correlated with low KL divergence.

\paragraph{KL regularization is not critical when using LoRA.} While previous work~\citep{ouyang2022training,stiennon2022learning} include the KL regularization penalty during PPO training, it is not necessary to achieve a high win rate within our LoRA-based experimental setup. In fact, as shown in Table~\ref{tab:kl_terms} and Figure~\ref{fig:win_rates}, a LoRA configuration without any KL regularization outperforms the released AlpacaFarm PPO checkpoint (48.2 vs 46.7). We conjecture that LoRA provides implicit regularization by freezing most of the parameters (e.g., feed-forward 
 layers, layernorm, and embeddings), which already discourages large deviations from the reference policy in parameter space. Previous work on other parameter-efficient fine-tuning methods~\citep{houlsby2019parameter} such as prompt tuning~\citep{lester2021power,vu-etal-2022-overcoming} and prefix tuning~\citep{li2021prefix} also demonstrate less overfitting in low-data regimes, which corroborates our hypothesis.
 One major caveat in our experiments is that properly verifying the regularization effect of LoRA requires a comparison to full-model fine-tuning without the KL penalty, for which we do not have adequate resources. 

\paragraph{Other divergence estimators outperform the standard KL estimator.}  Table~\ref{tab:kl_terms} shows that the KL estimator used in  open-source RLHF implementations such as TRLX underperforms alternative estimators. KL estimator does benefit from clamping (46.7 vs. 47.5 when clamped), but it is still worse than Jensen-Shannon regularizer (46.7 vs. 49.8) at step 100. The Jensen-Shannon estimator consistently outperforms all other estimators after step 60 in our experiments, and thus appears to be a promising alternative for future RLHF work. This finding aligns with previous work~\citep{go2023aligning}, who also show that the Jensen-Shannon divergence outperforms other divergence measures when fine-tuning language models to approximate energy-based models. 

\begin{figure*}
    \centering
    \includegraphics[width=0.495\textwidth]{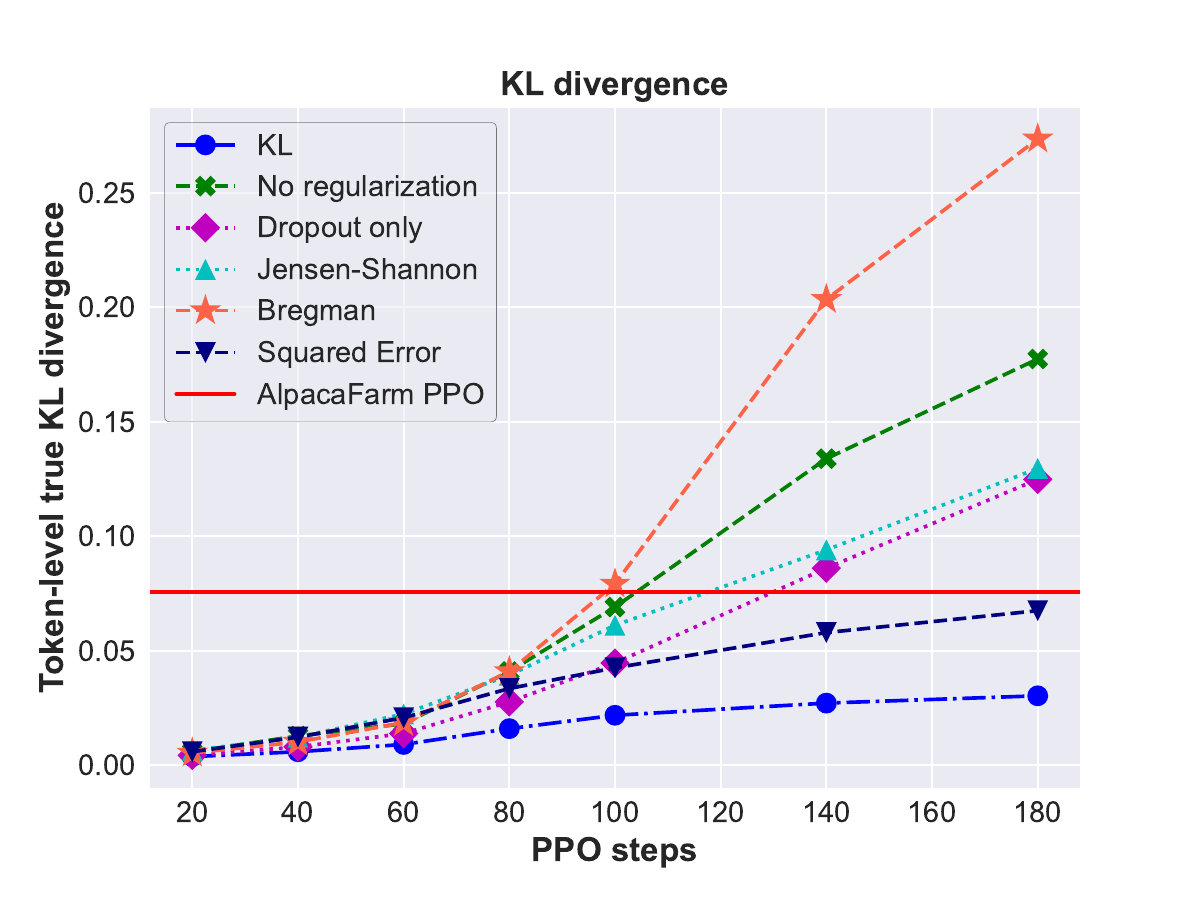}
    \includegraphics[width=0.495\textwidth]{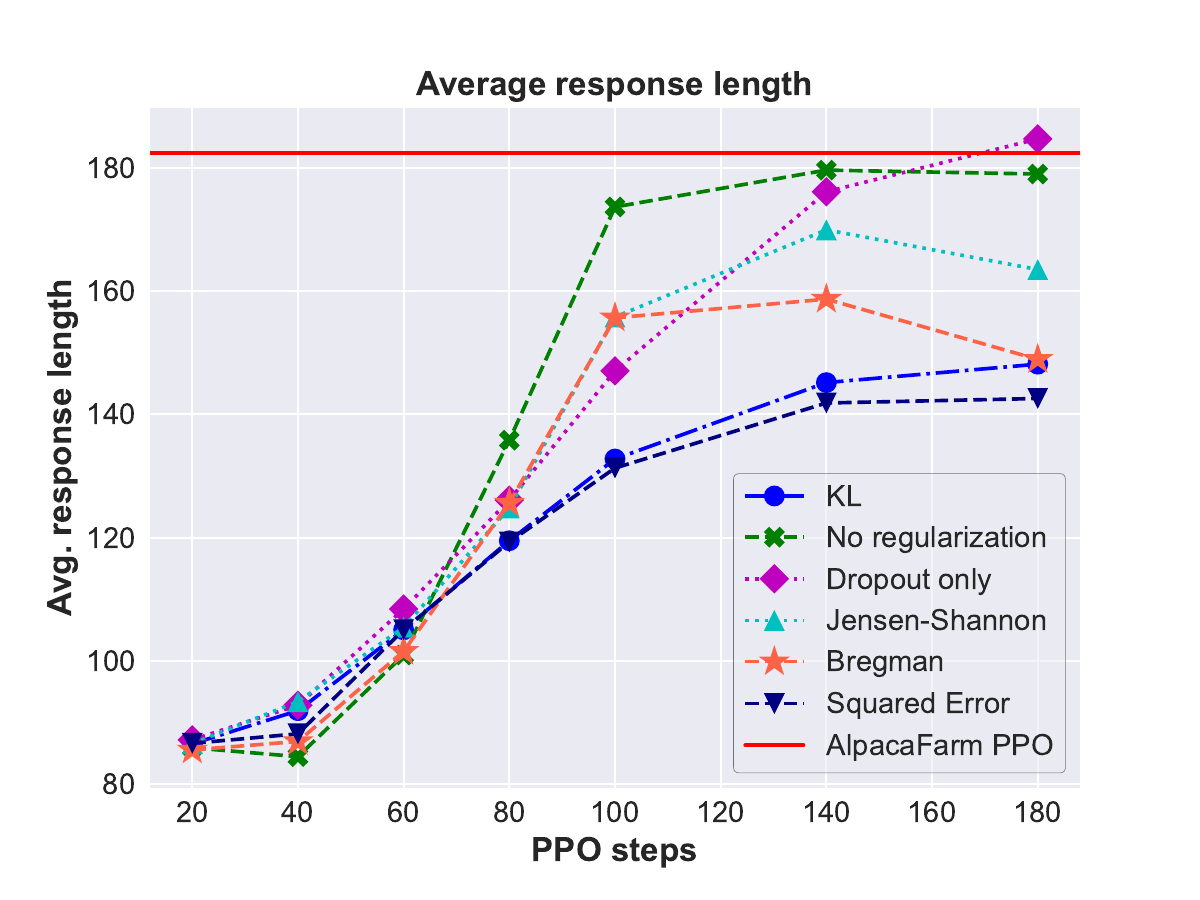}
    \caption{(\textbf{Left}): Regardless of the estimator, the true KL divergence (measured on 28K tokens sampled from AlpacaFarm) steadily increases as training proceeds. The standard KL estimator is predictably the most effective regularizer in terms of reducing true KL divergence. (\textbf{Right}): Response length increases until about 100 PPO steps, after which it plateaus or drops for most configurations.}
    \label{fig:true_kl}
\end{figure*}

\begin{figure*}
    \centering
    \includegraphics[width=0.495\textwidth]{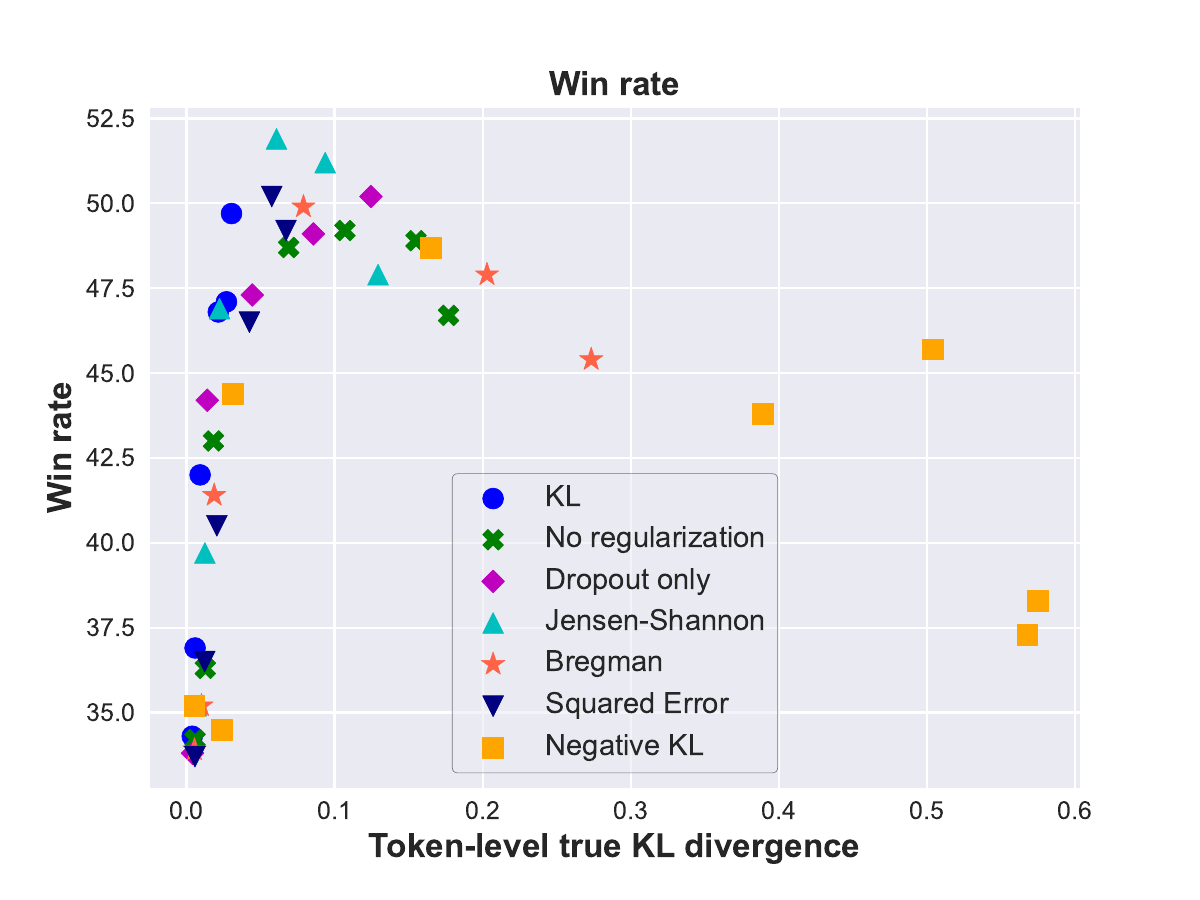}\includegraphics[width=0.495\textwidth]{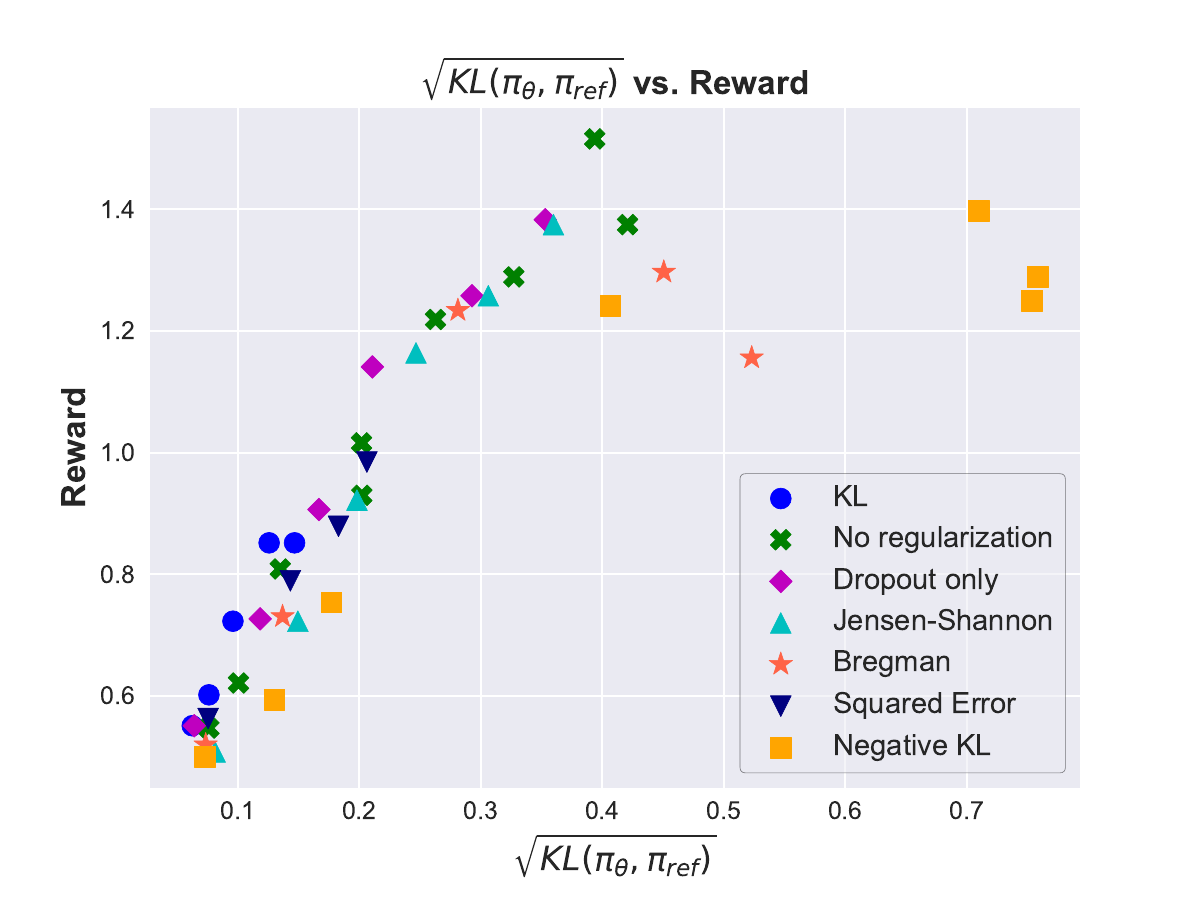}
    \caption{(\textbf{Left}): We plot the win rate vs. token-level true KL divergence for the checkpoints from steps 20, 40, 60, 80, 100, 140, and 180 of multiple configurations and multiple runs. In general win rates sharply increases and then more gradually decreases as the true KL divergence increases. (\textbf{Right}): The linear relationship between $\sqrt{\text{KL}(\pi_{\theta}, \pi_{\text{ref}})}$ and reward, observed in prior work, also holds in our LoRA implementation within a certain regime.}
    \label{fig:sqrt_kl}
\end{figure*}

\paragraph{We observe high win rates even when the KL divergence is moderately large.}  RLHF uses Monte-Carlo estimates of the KL divergence, and it is unclear how good these estimates are of the actual KL divergence (i.e., when computed over the entire distribution). To understand the effectiveness of minimizing the \emph{true} KL, we sample 28K tokens from the AlpacaFarm evaluation data and plot the KL divergence in Figure~\ref{fig:true_kl} (left). Regardless of the estimator used, the KL divergence increases as the training proceeds. The choice of regularizer impacts the speed in which KL grows: KL divergence increases faster without any regularization (``No regularization'') or when using the Bregman divergence,\footnote{In the Bregman divergence penalty, $\frac{\pi_{\text{ref}}(y)}{\pi_{\theta}(y)}$ is minimized, which encourages $\pi_{\theta}(y)$  to be large when $\pi_{\text{ref}(y)}$ is small despite the entropy bonus.} it grows slower when regularized by the standard KL and the squared error estimators. While removing KL penalty entirely leads to larger KL (e.g., 6 times that of the standard KL estimator at step 180), 
the resulting model still reaches win rates of $47\%\sim 49\%$, which are higher than that of the released AlpacaFarm PPO checkpoint.
That said, the best win rates are achieved when the KL divergence from the reference policy is neither small nor large (on our evaluation set, around $0.05$ to $0.12$ per token). 

\paragraph{A linear relationship between $\sqrt{\text{KL}(\pi_{\theta}, \pi_{\text{ref}})}$ and reward exists in our LoRA setup.}  Previous work~\citep{bai2022training,pmlr-v202-gao23h} demonstrates an approximately linear relationship between $\sqrt{\text{KL}(\pi_{\theta}, \pi_{\text{ref}})}$ and reward, which suggests that $\pi_{\theta}$ stays within a small region relative to $\pi_{\text{ref}}$ (i.e., $\pi_{\text{ref}} + \delta\pi_{\text{ref}}$) during PPO training.
In Figure~\ref{fig:sqrt_kl} (right), we confirm that this linear relationship also holds in our LoRA setup. To go beyond the $\delta\pi_{\text{ref}}$ region, we also experiment with a configuration that \emph{maximizes} KL during training. We find that this ``negative KL'' estimator leads to a region where the linear relationship breaks ($\sqrt{\text{KL}(\pi_{\theta}, \pi_{\text{ref}})}$ > 0.4 in our experiments). In this region, reward either plateaus or starts decreasing instead of linearly increasing. This suggests that the ratio between $\sqrt{\text{KL}(\pi_{\theta}, \pi_{\text{ref}})}$ and the reward can be a useful metric to monitor for over-optimization during training.
% \micomment{describe what happens in this region, plateau or steep drop in reward?}

\paragraph{We observe positive correlations between rewards, win rates, response length, and KL in certain regimes.} In addition to rewards, we find that win rates and response length also positively correlate with KL within a certain regime (step < 100 and KL < 0.12 in our experiments). Going beyond this region,  the average response length does not change much as KL divergence keeps increasing (Figure~\ref{fig:true_kl}, right), whereas win rates start to drop significantly (Figure~\ref{fig:sqrt_kl}, left). This has practical implications on PPO training -- early stopping leads to better performance while also saving compute.

\paragraph{PPO training has a larger negative impact on factuality with full model fine-tuning than with LoRA.} To quantify the effects of PPO training on the factuality of LLM-generated text, we evaluate several checkpoints using the FActScore metric~\citep{factscore}. FActScore evaluates the factual precision of a language model by breaking a long-form response into a collection of atomic facts and then computing the precision of these atomic facts.\footnote{We use the ChatGPT+retrieval configuration of FActScore. The prompt for all of our experiments is ``Tell me a bio of X'', and we perform all evaluations on the labeled split of 183 people entities released by ~\citet{factscore}.} Table~\ref{tab:factscore} shows that the SFT10k model (without any PPO training) obtains a higher FActScore than any checkpoint trained with PPO. Meanwhile, the released AlpacaFarm PPO checkpoint with full-model fine-tuning achieves the lowest FActScore (34.5\%), with all of our LoRA-based implementations outperforming it (39.4\% for the most comparable configuration). This result suggests that while PPO training can effectively steer the output to those preferred by humans for stylistic reasons (e.g., by increasing response length), it also may hurt the factuality of the generated text, and perhaps LoRA's regularization properties mitigates this effect to some extent. We provide example output on FActScore in Table~\ref{tab:factscore_example} and example output on AlpacaFarm in Table~\ref{tab:alpaca_example_1} and Table~\ref{tab:alpaca_example_2} in the Appendix.

\begin{table}[]
    \centering
    \begin{tabular}{lccc}
    \toprule
       \textbf{Configuration}  & \textbf{FActScore} ($\uparrow$) & \textbf{\# facts per response} &\textbf{\# tokens per response} \\ \midrule
       \multicolumn{4}{l}{\footnotesize{\emph{Publicly-released AlpacaFarm checkpoints
       from~\citet{dubois2023alpacafarm}}}}\vspace{0.1cm} \\
       AlpacaFarm SFT10k  & 39.7\% & 19.3 & 121.1\\
       AlpacaFarm PPO & 34.5\% & 37.1 & 247.9\\ \midrule
       \multicolumn{4}{l}{\footnotesize{\emph{Our PPO models trained with LoRA}}}\vspace{0.1cm} \\
       LoRA PPO w/ KL & 39.4\% & 26.7 & 170.8\\
       LoRA PPO w/ Jensen-Shannon & 38.2\% & 30.7 &  199.1\\
       LoRA PPO w/o regularization & 38.4\% & 33.9 & 217.5\\
    \bottomrule\vspace{0.1cm}
    \end{tabular}
    \caption{Evaluation on the FActScore labeled split~\citep{factscore}, which requires each model to generate a biography of 183 people entities. Models fine-tuned with the PPO training objective consistently underperform the SFT10k checkpoint in terms of factual precision. Implementing PPO training with LoRA somewhat mitigates the negative impact on factuality of model-generated text.}
    \label{tab:factscore}
\end{table}

\section{Related work}

Our work tackles the task of aligning large language models (LLMs) to human intents via instruction-following methods~\citep{ouyang2022training}. Instruction-following methods can be broadly categorized into (1) those that just perform supervised fine-tuning and (2) those that additionally apply reinforcement learning.

Methods in the first category differ in how they source instruction-following data for supervised fine-tuning. The instruction-response pairs can be generated from existing (close-sourced) models~\citep{,alpaca,honovich2022unnatural,mukherjee2023orca,wang-etal-2023-self-instruct}, curated by humans~\citep{köpf2023openassistant,zhou2023lima},  or even curated by LLMs~\citep{li2023selfalignment}. Open-source models, when fine-tuned on high quality~\citep{zhou2023lima} and diverse data~\citep{wang2023far}, can compete with blackbox LLMs on certain evaluations. 

RL-based methods~\citep{ziegler2020finetuning,ouyang2022training} align LLMs via online policy rollout and optimization. In contrast to SFT-based methods, RL incorporates feedback collected from either humans~\citep{stiennon2022learning,ouyang2022training} or AI~\citep{bai2022constitutional,lee2023rlaif}. These judgments are distilled into a preference (reward) model for evaluating responses during online policy rollout with (sparse) rewards~\citep{wu2023finegrained}, and the whole process is more involved than the  single-stage teacher-forced training used in SFT. While most RL-based methods depend on pairwise preference judgments~\citep{dubois2023alpacafarm}, feedback can take other forms including natural language~\citep{saunders2022selfcritiquing,fernandes2023bridging}. Recently,~\citet{bansal2023peering} show that human feedback protocol (e.g., rating or ranking) has significant impact on the evaluation of aligned LLMs.  

RL-based methods are typically regularized by a distributional term (e.g., a KL divergence penalty) to avoid degeneration caused by large deviations from a reference model. Previous work~\citep{korbak-etal-2022-rl} shows that KL-regularized RL can be viewed as Bayesian inference. Minimizing KL divergence is also related to distribution matching (DM) methods~\citep{CNTRL_NLG_ICLR2021}, where the target optimal distribution is available. Recently, ~\citet{korbak2022on} introduced KL-regularized RL from the perspective of DM, and ~\citet{go2023aligning} propose a framework that unifies KL-regularized RL, DM, and other $f$-divergence minimization methods.

\section{Conclusion}

In this technical report, we conduct an empirical analysis of the last stage of RLHF (PPO training) when implemented with low-rank adaptation, a parameter-efficient fine-tuning method. Our LoRA-based implementation of RLHF reduces the required hardware from eight to two A100 GPUs. Besides reduced memory consumption, we also observe that LoRA provides an implicit regularization effect during PPO training -- good performance can be achieved even when the KL regularization term in the PPO objective is removed. Additionally, we find that alternative regularizers to the standard KL divergence estimator (e.g., Jensen-Shannon divergence) lead to higher win rates. We release our code and LoRA weights to spur future research on RLHF from under-resourced labs.

\section*{Acknowledgements}
We would like to thank the AlpacaFarm authors for open-sourcing their code and models. We would like to thank Hiteshi Sharma, Banghua Zhu, Yang Liu, and Shi Dong for helpful discussions. This project was partially supported by awards IIS-1955567 and IIS-2046248 from the National Science Foundation (NSF).

\bibliography{anthology,custom}
\bibliographystyle{acl_natbib}

\appendix
\begin{table*}[]
    \centering
    \begin{tabular}{p{0.99\linewidth}}
        \toprule
        \textbf{Instruction} \hspace{1em} Tell me a bio of Don Featherstone (artist) \\ 
        \vspace{0.1em} \textbf{AlpacaFarm SFT10k} \hspace{1em} Don Featherstone (1935-2015) was an American artist and sculptor best known for creating the iconic plastic pink flamingo lawn ornament. He was born in 1935 in Hartford, Connecticut and studied at the Art Students League of New York and the Yale School of Art and Architecture. He worked as an artist and sculptor for the Garden Artists Guild in New York City, creating sculptures for the Smithsonian Institution and the Metropolitan Museum of Art. In 1957, he created the plastic pink flamingo lawn ornament for the Guild, which quickly became a popular icon of the mid-20th century American suburban landscape. He continued to create sculptures and artworks throughout his life, and his work is included in the collections of the Smithsonian Institution and the Metropolitan Museum of Art. He passed away in 2015 at the age of 80. \\
        \vspace{0.1em} \textbf{AlpacaFarm PPO} \hspace{1em} Don Featherstone (1921-1991) was an American artist and sculptor best known for creating the iconic plastic pink flamingo lawn ornament. He was born in 1921 in Los Angeles, California and studied art at the Chouinard Art Institute. He began his career as an artist and sculptor in the 1940s, creating sculptures out of clay and other materials. In the 1950s, he moved to Florida and began creating sculptures out of plaster. He was inspired to create the iconic pink flamingo lawn ornament in 1957, when he saw a flock of real pink flamingos in a Florida park. He created the first prototype out of plastic and presented it to the public the following year. The pink flamingo lawn ornament was an instant success and is now considered to be an iconic symbol of the 1950s. Don Featherstone continued to create sculptures out of plaster and other materials, but he is best known for his iconic pink flamingo lawn ornament. He passed away in 1991 in Florida. \\ 
        \vspace{0.1em} \textbf{AlpacaFarm PPO (LoRA)} \hspace{1em} Don Featherstone (1935-1991) was an American artist and sculptor best known for creating the iconic plastic pink flamingo lawn ornament. He was born in 1935 in Los Angeles, California and studied art at the Chouinard Art Institute. He worked as an artist and sculptor for several companies, including Union Products and Union Plastics, before founding his own company, Don Featherstone Studios, in 1960. He is best known for creating the pink flamingo lawn ornament, which he designed in 1957 and which became a popular fixture in suburban yards in the 1960s. He also designed several other garden ornaments, including the purple swan and the blue jay. He passed away in 1991.\\
        \vspace{0.1em} \textbf{No regularization} \hspace{1em} Don Featherstone (1935-1991) was an American artist and sculptor best known for creating the iconic plastic pink flamingo lawn ornament. He was born in 1935 in Massachusetts and studied art at the Massachusetts College of Art and the School of the Museum of Fine Arts in Boston. He began his career as an illustrator and graphic designer, creating advertisements and logos for companies such as Coca-Cola and Ford. In the late 1950s, he moved to New York City and began creating sculptures out of plaster and fiberglass. In 1957, he created his first plastic pink flamingo lawn ornament for the Union Products company. The flamingo quickly became a pop culture icon and is now considered to be a symbol of mid-century modern design. Featherstone continued to create sculptures and other works of art throughout his life, and his pieces can be found in museums and private collections around the world. He passed away in 1991 at the age of 56. \\  \bottomrule
        
    \end{tabular}
    \caption{Example of FActScore output. The actual biography of Don Featherstone can be found at \url{https://en.wikipedia.org/wiki/Don_Featherstone_(artist)}. Each of the responses contains hallucination to various degree. Models aligned with PPO training are less factually precise than SFT10K model.}
    \label{tab:factscore_example}
\end{table*}
\begin{table*}[]
    \centering
    \begin{tabular}{p{0.99\linewidth}}
        \toprule
        \textbf{Instruction} \hspace{1em} Verify the correctness of the given statement. ``For all integers j and k, if j and k are odd, then jk is odd.'' \\ 
        \vspace{0.1em} \textbf{AlpacaFarm SFT10k} \hspace{1em} True. \\
        \vspace{0.1em} \textbf{AlpacaFarm PPO} \hspace{1em} This statement is correct. For any two integers j and k, if both j and k are odd, then their product jk will also be odd. This is because the only even numbers that can be multiplied together to produce an odd number are 2 and 1. Since j and k are both odd, they can only multiply together to produce an odd number, and jk will also be odd. \\ 
        \vspace{0.1em} \textbf{AlpacaFarm PPO (LoRA)} \hspace{1em} This statement is correct. For any integers j and k, if j and k are odd, then jk is also odd. This is because the product of two odd numbers is always odd. Therefore, the statement is correct. \\
        \vspace{0.1em} \textbf{No regularization} \hspace{1em} This statement is correct. To see this, let's consider the cases where j and k are both even and where j and k are both odd. When j and k are both even, jk = (j - k) + (k - j) is even since both j - k and k - j are even numbers. When j and k are both odd, jk = (j - k) + (k - j) is odd since the sum of two odd numbers is odd. Therefore, the statement is correct. \\  \bottomrule
        
    \end{tabular}
    \caption{Example of AlpacaFarm output.}
    \label{tab:alpaca_example_1}
\end{table*}
\begin{table*}[]
    \centering
    \begin{tabular}{p{0.99\linewidth}}
        \toprule
        \textbf{Instruction} \hspace{1em} In a world where automation is becoming increasingly prevalent, is it more important to prioritize job creation or technological progress?  \\ 
        \vspace{0.1em} \textbf{AlpacaFarm SFT10k} \hspace{1em} It is more important to prioritize job creation in the short-term, as automation can lead to job losses and disruption of the existing economic order. In the long-term, however, technological progress should be prioritized in order to remain competitive in the global marketplace. \\
        \vspace{0.1em} \textbf{AlpacaFarm PPO} \hspace{1em} In my opinion, it is more important to prioritize job creation in a world where automation is becoming increasingly prevalent. While technological progress is undoubtedly important, it is essential to ensure that automation does not lead to mass unemployment. Job creation should be prioritized in order to ensure that people have access to meaningful employment, as well as the income and security that comes with it. Additionally, job creation should be prioritized in order to ensure that automation does not lead to a widening of economic inequality, as automation has the potential to disproportionately benefit those who are already economically advantaged. Job creation should also be prioritized in order to ensure that automation does not lead to a decrease in human autonomy, as automation has the potential to reduce the need for human labor and therefore reduce human agency. Therefore, I believe that job creation should be prioritized in a world where automation is becoming increasingly prevalent, in order to ensure that automation does not lead to negative consequences such as mass unemployment, widening economic inequality, and a decrease in human autonomy. \\ 
        \vspace{0.1em} \textbf{AlpacaFarm PPO (LoRA)} \hspace{1em} In my opinion, it is more important to prioritize job creation in the world where automation is becoming increasingly prevalent. Automation can lead to technological progress, but it can also displace workers and cause economic disruption. Therefore, it is essential to ensure that workers have access to meaningful employment and that the economy continues to grow and create new jobs. By prioritizing job creation, governments and organizations can help workers adjust to the changing job market and ensure that everyone has access to the opportunities and benefits of technological progress. \\
        \vspace{0.1em} \textbf{No regularization} \hspace{1em} In today's world, it is more important to prioritize job creation. Automation can create efficiencies and reduce costs, but it can also displace workers and lead to job losses. Therefore, it is essential to focus on creating new job opportunities and providing people with the skills they need to succeed in the labor market. This will ensure that people have access to meaningful employment and can contribute to their communities. At the same time, it is important to encourage technological progress and innovation, as this can lead to new job creation and improved standards of living. By balancing job creation and technological progress, governments can ensure that their citizens are able to benefit from the changing economy. \\  \bottomrule
        
    \end{tabular}
    \caption{Example of AlpacaFarm output.}
    \label{tab:alpaca_example_2}
\end{table*}

\end{document}